\documentclass[a4paper,10pt]{article}
\title{An Introduction To The Monogenic Signal}
\author{Christopher P.\ Bridge}

\usepackage{amsmath,graphicx}
\usepackage{amssymb} 
\usepackage{epstopdf} 
\usepackage{fullpage} 
\usepackage{tikz} 
\usepackage{3dplot} 
\usepackage{caption,subcaption} 
\usepackage{url} 

\DeclareMathOperator{\sgn}{sgn}
\newcommand*\conj[1]{\overline{#1}} 
\newcommand{\imagi}{i} 

\begin{document}
\maketitle

\section{Introduction and Overview}
\label{sec:intro}

The monogenic signal is an image analysis methodology that was introduced by Felsberg and Sommer in 2001 and has been employed for a variety of purposes in image processing and computer vision research. In particular, it has been found to be useful in the analysis of ultrasound imagery in several research scenarios mostly in work done within the BioMedIA lab at Oxford.

However, the literature on the monogenic signal can be difficult to penetrate due to the lack of a single resource to explain the various principles from basics. The purpose of this document is therefore to introduce the principles, purpose, applications, and limitations of the methodology. It assumes some background knowledge from the fields of image and signal processing, in particular a good knowledge of Fourier transforms as applied to signals and images. We will not attempt to provide a thorough mathematical description or derivation of the monogenic signal, but rather focus on developing an intuition for understanding and using the methodology and refer the reader elsewhere for a more mathematical treatment.

My MATLAB/GNU Octave implementation of the monogenic signal (used for figures in this document) is available on Github under the GNU Public License at \url{https://github.com/CPBridge/monogenic_signal_matlab}.

\section{1D Signals and the Analytic Signal}
\label{sec:analytic}

The \emph{analytic signal} \cite{Stanford} is a widely-used concept in signal processing concerning the analysis of 1D signals. It is essentially an alternative representation of any real-valued signal using a complex-valued signal that makes various processes (such as modulation and demodulation) conceptually simpler. It is inherently a 1D concept and therefore has no place in image analysis\footnote{The analytic signal is in fact vital to the creation of ultrasound images from the reflected radio-frequency signals that are picked up by the transducer as it is used to recover the amplitude envelope. Several such demodulated signals from different scans lines are then combined to form a B-mode image. This stage of processing is, however, not the focus of the present document as we assume that all images have already gone through the standard processing chain to give human-viewable images.} but we will discuss the analytic signal because the concepts introduced are vital to understanding the monogenic signal, which is a generalisation of the analytic signal to multidimensional signals such as images.

Recall that the Fourier transform, $F(\omega)$, of any real-valued signal, $f(t)$, displays \emph{Hermitian symmetry}; that is it has components at negative frequencies that are the complex conjugate of the positive frequency components:

\begin{equation}
	F(-\omega) = \conj{F(\omega)} ,
\end{equation}

\noindent where $\conj{x}$ denotes the complex conjugation operation on $x$. The analytic signal is a representation of the our original signal, $f(t)$, using \emph{only the components at positive frequencies}: we can simply discard the negative frequency components without loss of information because there is redundancy in the Fourier transform of a real-valued signal due to the Hermitian symmetry. Clearly, however, the Hermitian symmetry of the Fourier transform is broken by this operation, meaning that the resulting time-domain analytic signal is no longer real-valued, but complex-valued. We will denote the spectrum of this analytic signal as $F_a(\omega)$, given by:

\begin{equation}
	\label{eqn:analyticases}
	F_a(\omega) = \begin{cases}
					2 F(\omega), & \omega > 0 \\
					0, & \omega < 0 \\
					F(0), & \omega = 0 .
              	 \end{cases}
\end{equation}

Notice that we double all the positive frequency components to keep the same power of the resulting signal, and leave the zero-frequency (DC) component unchanged.

\begin{figure}
	\centering
	\includegraphics[width=\textwidth]{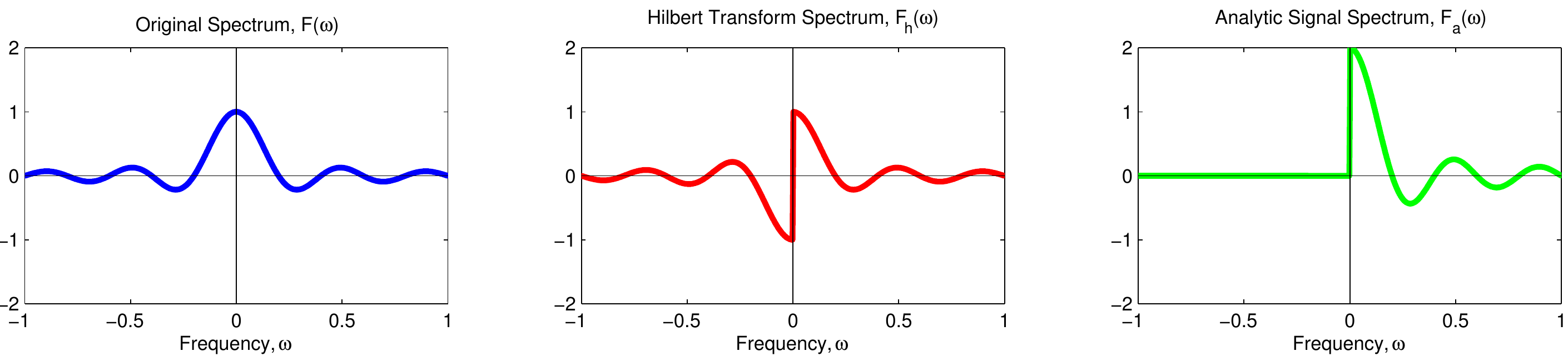}
	\caption{Formation of the analytic signal in the frequency domain (only real parts shown): the spectrum of the original, real signal (\textit{left}) displays conjugate symmetry. The Hilbert transform of the original signal is odd symmetric (\textit{centre}). Adding the two gives the spectrum of the analytic signal (\textit{right}), which has no negative-frequency components.}
	\label{fig:spectra}
\end{figure}

Conceptually, we can think of the operation of removing the negative frequency components from the original spectrum, $F(\omega)$, as adding a carefully constructed odd symmetric (anti-symmetric) spectrum to the spectrum of the original signal (see Figure \ref{fig:spectra}).

\begin{equation}
	\label{eqn:analyticadding}
	F_a(\omega) = F(\omega) + F_h(\omega) ,
\end{equation}

\noindent where $F_h(\omega)$ is formed by creating an `odd symmetric version' of $F(\omega)$ as follows:

\begin{equation}
	\label{eqn:hilbertcases}
	F_h(\omega) = \begin{cases}
					F(\omega) & \omega > 0 \\
					-F(\omega) & \omega < 0 \\
					0 & \omega = 0 ,
				 \end{cases}
\end{equation}

\noindent or, equivalently, we can write using the \emph{signum} function ($\sgn(\cdot)$):

\begin{equation}
	\label{eqn:hilbertsgn}
	F_h(\omega) = F(\omega) \cdot \sgn(\omega)
\end{equation}

\noindent where

\begin{equation}
\sgn(x) = \begin{cases}
             1, & x > 0 \\
             -1, & x < 0 \\
             0, & x = 0 .
		  \end{cases}
\end{equation}

In this way we can also rewrite the relationship between a signal and its analytic signal (Equation (\ref{eqn:analyticases})) as simple multiplication in the frequency domain:

\begin{equation}
	\label{eqn:analyticmultiplier}
	F_a(\omega) = \left(1+\sgn(\omega)\right) F(\omega) ,
\end{equation}

\begin{figure}
	\centering
	\includegraphics[width=\textwidth]{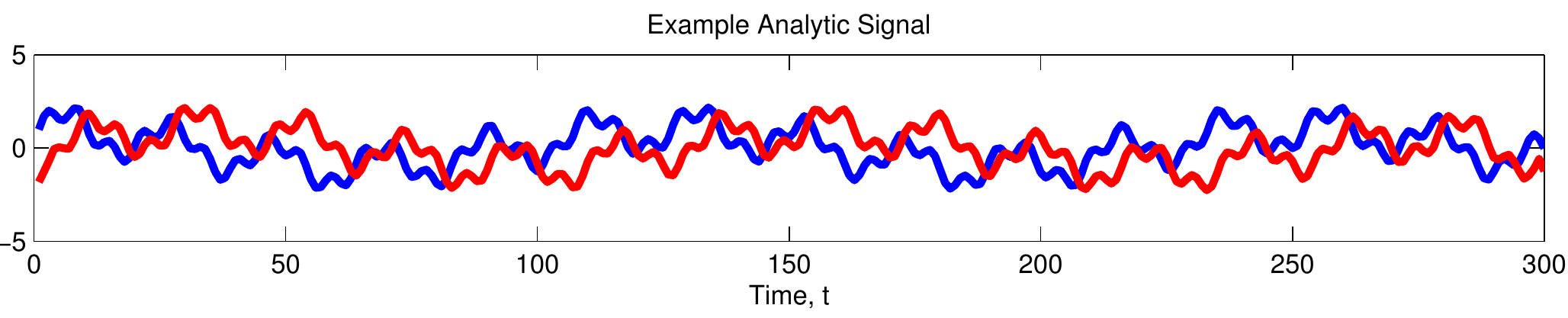}
	\caption{Example of an analytic signal. The analytic signal has a real part (\textit{blue}) that is the same as the original signal, and an imaginary part (\textit{red}) formed from the Hilbert transform of the original signal}
	\label{fig:analytic}
\end{figure}

Returning to Equation (\ref{eqn:analyticadding}), we have created the spectrum of the analytic signal, $F_a(\omega)$, by adding two spectra in the frequency domain. As we have discussed, the original spectrum, $F(\omega)$, is symmetric and therefore represents a \emph{purely real} time-domain signal. Furthermore, the spectrum $F_h(\omega)$ is odd symmetric by construction and therefore represents a \emph{purely imaginary} time-domain signal, which we will call $f_h(t)$. Therefore, in the time domain, the real part of the analytic signal is simply the original signal, and we have added a new imaginary part that `suppresses' the negative frequency components (see Figure \ref{fig:analytic}):

\begin{equation}
	f_a(t) = \underbrace{f(t)}_{\text{purely real}} + \underbrace{f_h(t)}_{\text{purely imaginary}} .
\end{equation}

The process of forming the signal $f_h(t)$ from the original signal $f(t)$ is called the \emph{Hilbert transform} and $f_h(t)$ is called the \emph{Hilbert transform} of $f(t)$. The Hilbert transform may also be represented in the time domain by convolution with the distribution $\frac{1}{\pi t}$:

\begin{equation}
	f_h(t) = \imagi \, \cdot \, \text{p.v.} \int_{-\infty}^{\infty} \frac{f(t)}{\pi (t - \tau)} \, \mathrm{d}\tau ,
\end{equation}

\noindent where $\text{p.v.}$ is the Cauchy principal value of the integral and is necessary because of the singularity at $ t = \tau$. However, it is generally simpler both conceptually and computationally to use the frequency-domain definition of the Hilbert transform as described above.

Note that an alternative definition for the Hilbert transform is $F_h(\omega) = -\imagi F(\omega) \cdot \sgn(\omega)$, where the multiplication by $\imagi$ means that the resulting time-domain signal is purely real and correspondingly $F_a(\omega) = F(\omega) + \imagi F_h(\omega)$ and $f_a(t) = f(t) + \imagi f_h(t)$. This form is more useful when the Hilbert transform is considered in isolation, as the result of applying the transform to a real-value signal is a second real-valued signal.

In MATLAB, one can use the Signal Processing Toolbox function `hilbert' to work with the analytic signal. Despite its name, this function returns the complex-valued analytic signal -- the Hilbert transform can be found by then extracting the imaginary part of this using the `imag' function.

\subsection{Local Amplitude and Local Phase}
\label{ssec:phase}

Let us consider the analytic signal of a simple sinusoid of some fixed amplitude, $A$, and frequency, $\omega_0$:

\begin{equation}
f(t) = A \sin(\omega_0 t) .
\end{equation}

Applying the above relationships, we find that the Hilbert transform of $f(t)$ is a cosine wave, and the analytic signal is therefore a complex exponential.

\begin{align}
f_h(t) &= \imagi A \cos(\omega_0 t)\\
f_a(t) &= f(t) + f_h(t) \\
	   &= A\sin(\omega_0 t) + \imagi A \cos(\omega_0 t) \\
       &= Ae^{\imagi\omega_0 t} .
\end{align}

In other words, the Hilbert transform has manifested itself as a \emph{phase shift} of $-\frac{\pi}{2}$ (one quarter cycle) of the original signal (this is because a cosine wave is equivlent to a sine wave with as phase shift of $-\frac{\pi}{2}$). As Fourier theory tells us that a general signal (such as that in Figure \ref{fig:analytic}) may be written as a (possibly infinite) sum of scaled and shifted sinusoids, it is straightforward to see that the Hilbert transform phase shifts \emph{each} frequency component by $-\frac{\pi}{2}$. The signal $f(t)$ and its Hilbert transform $f_h(t)$ are said to be in \emph{phase quadrature}.

The analytic version of our sinusoidal signal is useful to us as it allows us to extract values for the amplitude, $A$, and phase, $\phi = \omega_0 t$, of the signal by expressing the complex analytic signal in polar form:

\begin{align}
	A(t) &= \sqrt{f(t)^2 + f_h(t)^2} \\
	\phi(t) &= \arctan{\left(\frac{f_h(t)}{f(t)}\right)} .
\end{align}

However this result is not limited to simple sinusoids of fixed amplitude and constant frequency, but can in fact be applied to general signals. We can consider the analytic form of a signal to be produced by a complex exponential with time-varying amplitude and frequency:

\begin{equation}
f_a(t) = A(t)e^{\imagi \phi (t)} .
\end{equation}

\begin{figure}
	\centering
	\includegraphics[width=\textwidth]{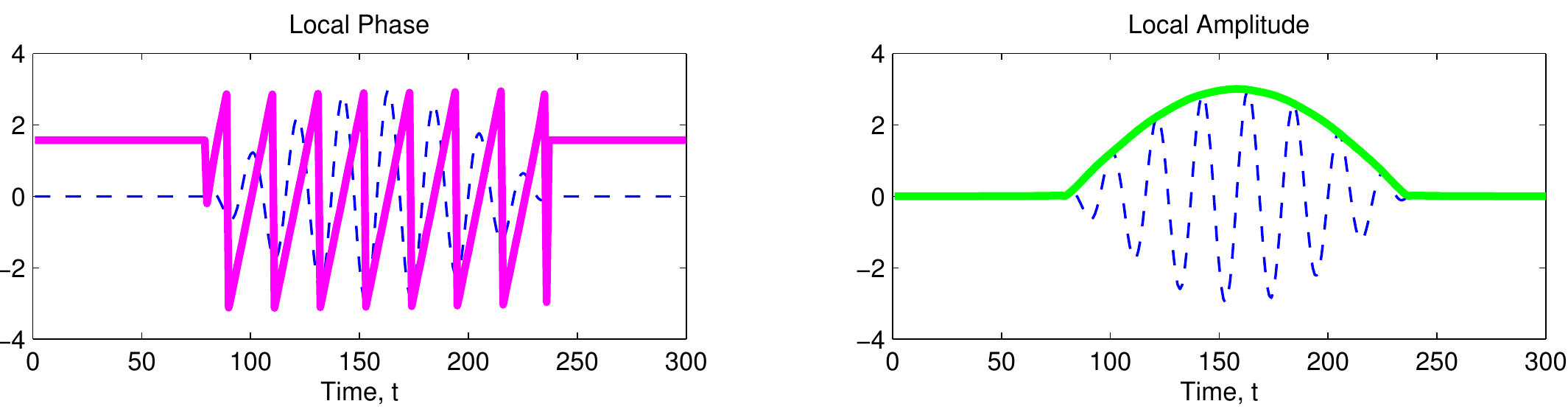}
	\caption{The local phase (\textit{left, magenta}) and local amplitude (\textit{right, green}) measures of a sine wave modulated by a cosine pulse (\textit{blue, dashed}). It is apparent that the local amplitude gives the envelope of the oscillations, while the local phase gives the point in the cycle of the local oscillations. }
	\label{fig:envelope}
\end{figure}

This gives rise to the concepts of \emph{local phase} and \emph{local amplitude}, which are the phase and amplitude of this polar representation. The local amplitude is a measure of the envelope of the signal at that point, and the local phase is a measure of the shape of the signal at that point (e.g.,\ a phase of $0$ corresponds to a `peak' part of a signal and a phase of $\pi$ corresponds to a `trough'), thus we have achieved a separation of these two important characteristics of a signal (some authors refer to this as the `split of identity'). We can also define \emph{local frequency} as the rate of change of local phase. Figure \ref{fig:envelope} shows the local phase and local amplitude measures of a modulated sinusoidal signal. It is easy to see from this example why the analytic signal is useful for signal processing operations such as demodulation and envelope detection.

\subsection{Scale in Local Phase Analysis}
\label{ssec:scale}

\begin{figure}
	\centering
	\includegraphics[width=\textwidth]{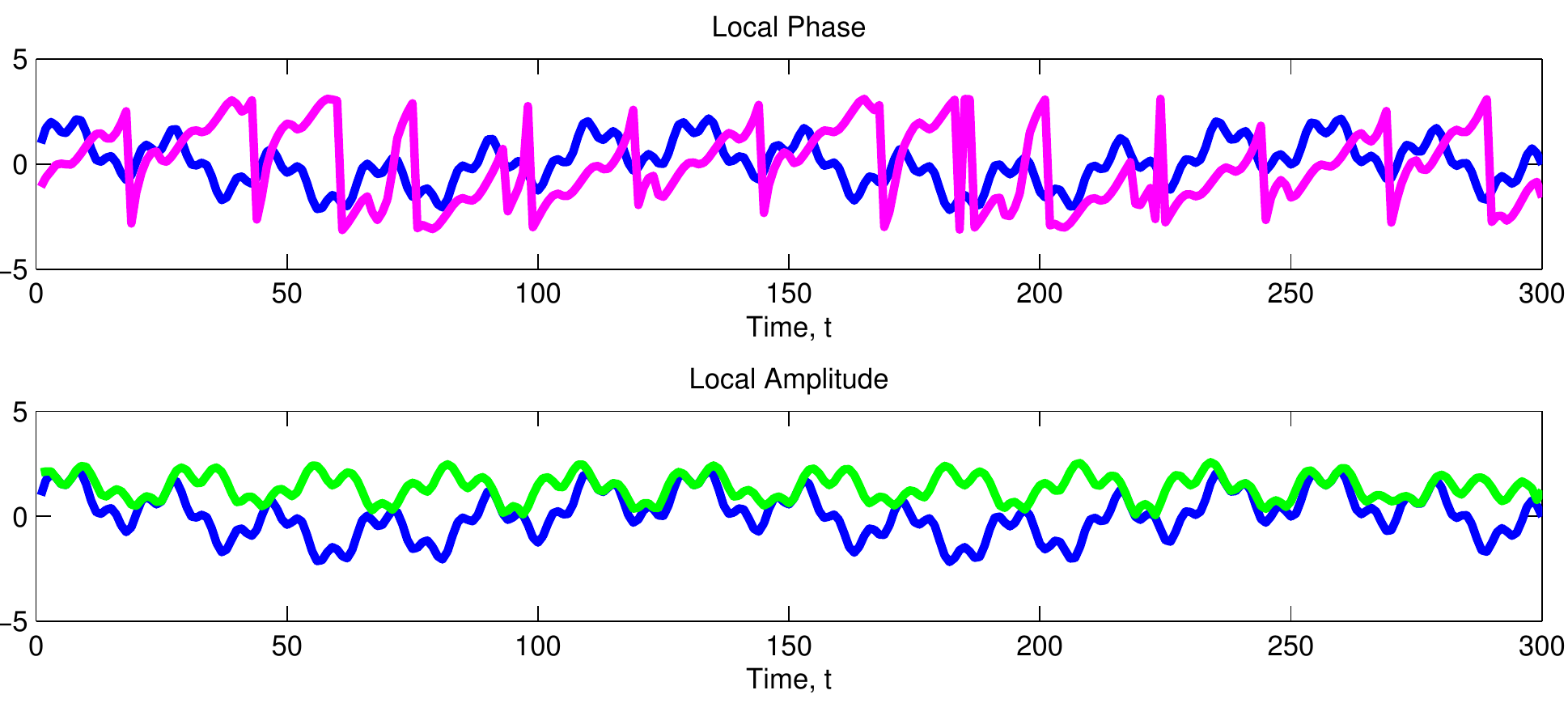}
	\caption{The local phase (\textit{top, magenta}) and local amplitude (\textit{bottom, green}) measures of the signal shown in Figure \ref{fig:analytic} (\textit{blue}). }
	\label{fig:unfiltered}
\end{figure}

Our example in Figure \ref{fig:envelope} demonstrated a very well-behaved analytic signal where the interpretation of the local phase and local amplitude was very clear. However, for more general signals this is not always the case. Consider instead the example in Figure \ref{fig:analytic}, which is composed of the sum of three sinusoidal oscillations at different frequencies. It is not clear in this case what constitutes the amplitude envelope, and what constitutes the oscillations within this envelope. Indeed we find that the local amplitude and local phase measures do not help us much in this regard (see Figure \ref{fig:unfiltered}).

The problem is that structure exists in this signal at different scales -- to get a useful output we need to choose what scale of structure we are interested in. We can do this by filtering the signal with a bandpass filter centred on the frequencies of interest before calculating the analytic signal. There are a number of different filter types that may be used for this purpose, for example Gabor, log-Gabor, Poisson, Cauchy, Gaussian derivative and others. We will focus on log-Gabor filters here as they are a popular choice.

A log-Gabor filter \cite{Field1987,KovesiWeb} (sometimes referred to as a log-normal filter) is defined in the frequency domain as having a Gaussian frequency response when viewed on a logarithmic frequency axis:

\begin{equation}
	\label{eqn:loggabor}
	G(\omega) = \exp\left( -\frac{\left(\log\left(\frac{\lvert\omega\rvert}{\omega_0}\right)\right)^2}{2\left(\log\left(\sigma_0\right)\right)^2}\right) .
\end{equation}

\begin{figure}
	\centering
	\includegraphics[width=\textwidth]{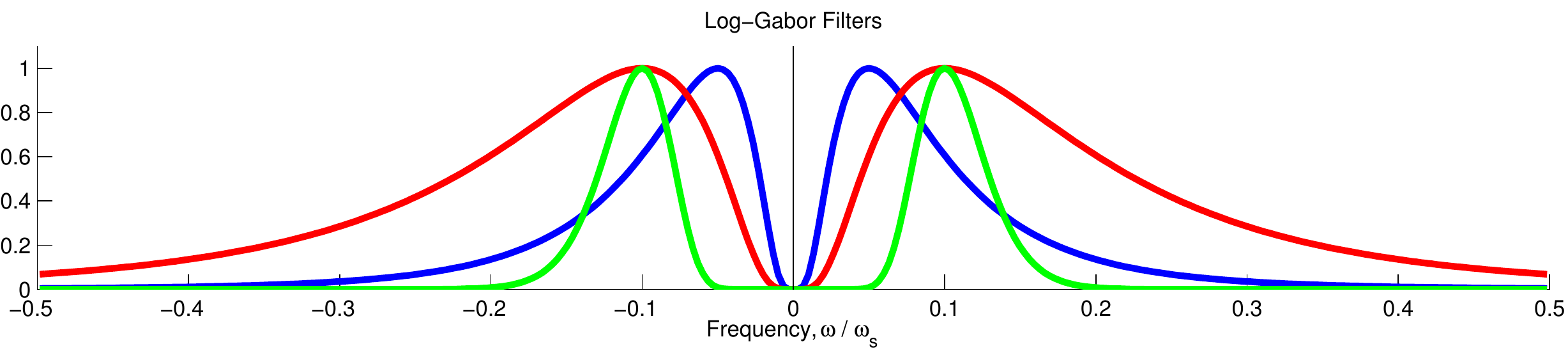}
	\caption{Frequency domain representations of log-Gabor filters. \textit{Red}: $\omega_0 = 0.2\omega_s$, $\sigma_0 = 0.5$;  \textit{Blue}: $\omega_0 = 0.1\omega_s$, $\sigma_0 = 0.5$; \textit{Green}: $\omega_0 = 0.2\omega_s$, $\sigma_0 = 0.8$ where $\omega_s$ is the sampling frequency}
	\label{fig:filters}
\end{figure}

Log-Gabor filters are usually defined in the frequency domain since there is no closed-form expression for their time-domain form (i.e.\ impulse response). There are two degrees of freedom when designing a log-Gabor filter. The first, $\omega_0$, is the centre-frequency of the passband. It governs what scale structures are picked out by the filter. Structures with wavelengths approximately equal to that of the filter, $\lambda_0 = 2\pi/\omega_0$, will be selected. The second, $\sigma_0$, is a shape parameter that governs how selective the filter is in the frequency domain (i.e.\ it governs the bandwidth of the passband). See Figure \ref{fig:filters} for some examples of log-Gabor filters. One of the key advantages of log-Gabor filters is that they can be constructed with arbitrary bandwidth (by varying $\sigma_0$) whilst maintaining a gain of zero at a $\omega = 0$ (the DC frequency); this can clearly be seen in the figure. Further discussion of the different filter types is out of the scope of this document, but see Boukerroui et al.~\cite{Boukerroui2004} for a detailed comparison.

In order to apply a log-Gabor filter (or other filter) to the problem of scale-selectivity with the analytic signal, we can alter Equation (\ref{eqn:analyticmultiplier}) to include a filtering stage before finding the analytic signal by multiplying by our frequency-domain filter, $G(\omega)$:

\begin{equation}
\label{eqn:analyticmultiplierwithfilter}
F_a(\omega) = \left(1 + \sgn(\omega)\right)G(\omega)F(\omega) .
\end{equation}

\begin{figure}
	\centering
	\includegraphics[width=\textwidth]{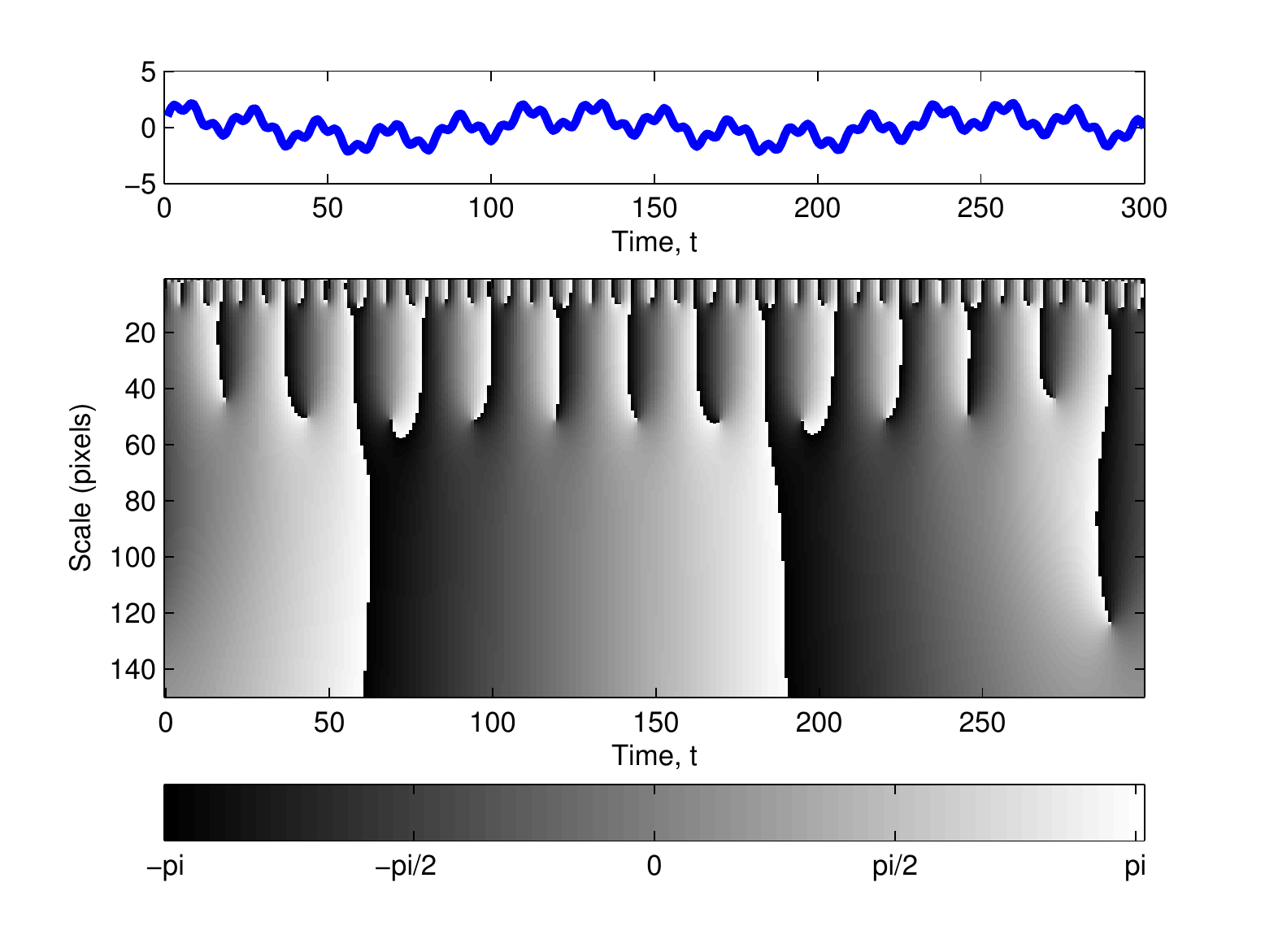}
	\caption{Scaleogram of local phase for the signal in Figure \ref{fig:analytic} (shown above). The image represents local phase of the signal through time along the horizontal axis and over different scales on the vertical axis. The grey value of each pixel represents the local phase value according to the scale below the image. At each different scale (horizontal line on the image) the local phase was calculated after applying a log-Gabor filter at the relevant scale (i.e.\ wavelength $\lambda_0$) and shape parameter $\sigma_0 = 0.6$. The different interpretations of the signal structure at different scales are evident.}
	\label{fig:scalogram}
\end{figure}

The `scaleogram' in Figure \ref{fig:scalogram} shows the scale selection behaviour of the local phase measure. At the smaller scales ($<10$ pixels), the local phase measure describes the very small scale oscillations, and then becomes dominated by the medium scale changes as the centre-wavelength of the filter increases, before finally becoming dominated by the long term trends at scales of about 60 pixels and greater.

It is important to realise how this local phase and amplitude model differs from the Fourier transform. The Fourier transform expresses the \emph{entire signal} as the sum of complex exponentials with different amplitudes and phases. It therefore provides localisation in frequency but has infinite spatial extent. By contrast, the local phase model that we have discussed is localised in space, as the phase and amplitude estimates are local, but only covers a range of frequencies due to the bandpass filtering. Therefore, the model can be seen as a compromise between localisation in space and in frequency. For further discussion of this idea, see the famous paper by Field \cite{Field1987}.

\subsection{Odd and Even Quadrature Filters}
\label{ssec:oddeven}

Take another look at Equation (\ref{eqn:analyticmultiplierwithfilter}). We described this process as first filtering the signal (i.e.\ finding $G(\omega)F(\omega)$) and then finding the analytic signal of this filtered signal using multiplication by $\left(1+\sgn(\omega)\right)$. However, there is a another equivalent and equally valid interpretation of Equation (\ref{eqn:analyticmultiplierwithfilter}): first find the analytic representation \emph{of the filter} by multiplying its spectrum in the usual way:

\begin{equation}
	G_a(\omega) = G(\omega) \left(1+\sgn(\omega)\right) ,
\end{equation}

\noindent and now use this new `analytic' filter to filter the original signal. If we choose to think along these lines, what can we say about our `analytic' filter? Just like any analytic signal, it is complex-valued with the real part being the original filter impulse response, and the imaginary part being in quadrature with it. This means that if we use the filter on a real signal, the result is a complex-valued signal where the real part is the signal filtered with the real part of the filter, and the imaginary part is the signal filtered with the imaginary part of the filter.

\begin{figure}
	\centering
	\includegraphics[width=\textwidth]{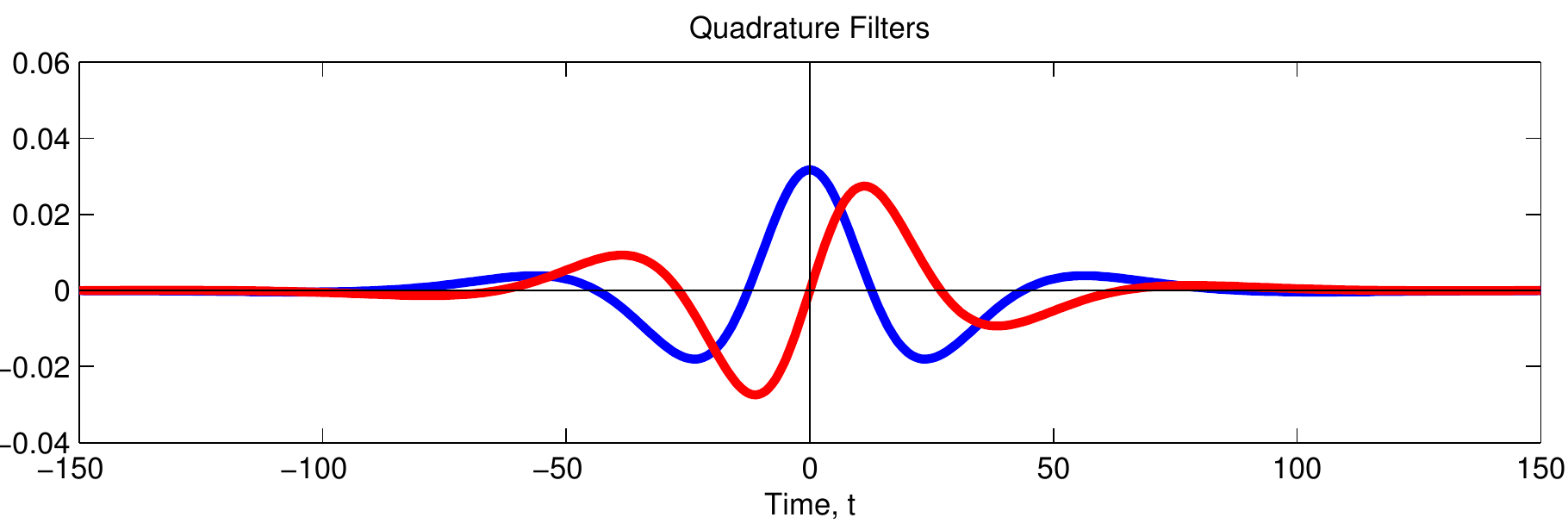}
	\caption{Time-domain impulse responses of a log-Gabor quadrature pair: \textit{blue} real, even-symmetric part, \textit{red} imaginary, odd-symmetric part. These have been calculated from the frequency-domain representation (Equation (\ref{eqn:loggabor})) using an inverse FFT, as there is no closed-form expression.}
	\label{fig:quadrature}
\end{figure}

Additionally, bandpass filters, such as the log-Gabor filter, have symmetric impulse responses and therefore real-valued symmetric spectra. Consequently, the Hilbert transform (imaginary part) of the filter has a real-valued, odd symmetric spectrum and an odd symmetric filter response. This is also evident from the phase-shift argument -- an even-symmetric function may be written as a sum of cosines and the Hilbert transform induces a phase shift of $-\pi/2$, meaning that the resulting signal may be written as a sum of pure sines and must therefore be odd-symmetric. See Figure \ref{fig:quadrature} for an illustration of a log-Gabor quadrature filter.

It is therefore common to think of the process of finding the analytic signal as convolving the original signal with two filters: one even-symmetric (or `even' filter) and one odd-symmetric (or `odd' filter). However it is important to note that, as we have just discussed, this is mathematically equivalent to finding the analytic representation of the signal after filtering by the even filter alone.

\subsection{Relation to Gradient}
\label{ssec:gradient}

We have said that the Hilbert transform of a signal can be thought of in terms of phase-shifting the frequency components in the signal. Are there any other interpretations? Look again at our definition of the Hilbert transform in the frequency domain, Equations (\ref{eqn:hilbertcases}) and (\ref{eqn:hilbertsgn}). We can rewrite these in yet another equivalent form (this time we multiply by $-\imagi$ to give a real time-domain signal):

\begin{equation}
	\label{eqn:hilbertfrac}
	F_h(\omega) = \begin{cases}
					-\imagi \frac{\omega}{\lvert \omega \rvert} \cdot F(\omega) , & \omega \neq 0 \\
					0, & \omega = 0 .
				\end{cases}
\end{equation}

\noindent Compare this to the Fourier domain form of the derivative operation on a signal:

\begin{equation}
	\label{eqn:fourierderiv}
	F'(\omega) = \imagi \omega F(\omega) ,
\end{equation}

\noindent where $F(\omega)$ is the Fourier transform of $f(t)$ and $F'(\omega)$ is the Fourier transform of $f'(t) = \frac{\mathrm{d}}{\mathrm{d}t}f(t)$. We can see that the Hilbert transform of a signal can be thought of as a (negative) smoothed gradient, where the smoothing kernel is $\lvert \omega \rvert ^{-1}$ in the frequency domain. In fact, the relationship between the gradient and the Hilbert transform is deeper than this and has its roots in fractional calculus. However for the purposes of this document we will simply note that it can be helpful to think of the Hilbert tranform as a smoothed gradient and refer the reader to Unser et al.~\cite{Unser2009} for more details.

\section{2D Images and the Monogenic Signal}
\label{sec:mono}

Given the usefulness of the analytic signal in signal processing and interpretation, it is natural to try to extend the approach to two dimensions. Indeed the importance of phase in images has long been established \cite{Oppenheim1981}. Ideally we would like a way to extract a phase quadrature image from an original image, such that we can extract local phase and amplitude information. Unfortunately, this process is not straightforward in two dimensions because we now need to consider different directions. Some authors have tried to generate representations that consider the Hilbert transform along a preferred or pre-selected direction in the image. However, we have only generated a truly useful representation if it is \emph{isotropic}, i.e.\ treats all directions in the image in the same way. The \emph{monogenic signal} of Felsberg and Sommer~\cite{Felsberg2000b,Felsberg2001} is the only isotropic generalisation of the analytic signal, and has found applications in a number of different image processing tasks. It is this approach that we will consider from now on (see \cite{Felsberg2001} for a discussion of other approaches).

Recall that to create the analytic signal we generated a single imaginary, phase-quadrature part via the Hilbert transform and used it in combination with the (real) original signal to give the analytic signal. Just like we need two components to express the gradient in an image, it turns out that in the 2D case we need \emph{two} imaginary parts -- one for each axis direction -- and therefore we \emph{cannot} represent the monogenic signal using a complex number. It is instead necessary to turn a branch of mathematics called \emph{geometric algebra}, and in particular to the algebra of \emph{quaternions}, which in simple terms can be thought of as generalised complex numbers with one real part and three imaginary parts. This makes the derivation and some of the principles of the monogenic signal rather inaccessible to those without the relevant mathematical background, and unfortunately geometric algebra is not well-known within the image processing community. This does not mean, however, that we cannot understand and use the monogenic signal. Imagine if we had presented \S\ref{sec:analytic} with no understanding of complex numbers: we could simply have treated the signal and its Hilbert transform as two different signals and neglected the fact that they in fact form the real and imaginary parts of a single complex signal. This is the approach we will take here for the monogenic signal: we shall (at least initially) present the monogenic signal as having three real-valued parts. However we should bear in mind that in fact we are really dealing with different parts of the same signal within the quaternion framework.

For obvious reasons, we will focus our discussion on the monogenic signal of 2D images, however everything here generalises trivially to higher dimensions by adding further odd parts (one for each dimension). Indeed previous work has shown the 3D monogenic signal to be useful in analysing 3D volumetric ultrasound data \cite{Rajpoot2009}.

To generate the two odd parts from the original signal, we need to use a generalisation of the Hilbert transform called the \emph{Riesz transform}. Let $f(\mathbf{x})$ be a 2D image of a spatial variable $\mathbf{x} = (x,y)^T$, and let $F(\boldsymbol{\omega})$ be its frequency-domain representation found using the 2D Fourier transform, where $\boldsymbol{\omega} = (\omega_x,\omega_y)^T$ is a two-dimensional frequency. The two odd parts of the monogenic signal, $F_{o1}(\boldsymbol{\omega})$ and $F_{o2}(\boldsymbol{\omega})$, are found in the frequency domain as follows:

\begin{align}
	\label{eqn:riesz1}
	F_{o1}(\boldsymbol{\omega}) &= \begin{cases} \imagi \frac{\omega_x}{\lVert \boldsymbol{\omega} \rVert}F(\boldsymbol{\omega}), & \boldsymbol{\omega} \neq 0 \\ 0, & \boldsymbol{\omega} = 0 , \end{cases} \\
	\label{eqn:riesz2}
	F_{o2}(\boldsymbol{\omega}) &= \begin{cases} \imagi \frac{\omega_y}{\lVert \boldsymbol{\omega} \rVert}F(\boldsymbol{\omega}), & \boldsymbol{\omega} \neq 0 \\ 0, & \boldsymbol{\omega} = 0 . \end{cases}
\end{align}

\noindent We can show using symmetry arguments\footnote{In 2D a real-valued image has a spectrum with an even-symmetric real part ($\operatorname{Re}\lbrace F(\omega_x,\omega_y)\rbrace = \operatorname{Re}\lbrace F(-\omega_x,-\omega_y)\rbrace$ and $\operatorname{Re}\lbrace F(-\omega_x,\omega_y)\rbrace = \operatorname{Re}\lbrace F(\omega_x,-\omega_y)\rbrace$) and an odd-symmetric imaginary part ($\operatorname{Im}\lbrace F(\omega_x,\omega_y)\rbrace = -\operatorname{Im}\lbrace F(-\omega_x,-\omega_y)\rbrace$ and $\operatorname{Im}\lbrace F(-\omega_x,\omega_y)\rbrace = -\operatorname{Im}\lbrace F(\omega_x,-\omega_y)\rbrace$). Since in Equations (\ref{eqn:riesz1}) and (\ref{eqn:riesz2}), $F(\boldsymbol{\omega})$ is the transform of a real image, it obeys the symmetry conditions. Following this through, $F_{o1}(\boldsymbol{\omega})$ and $F_{o1}(\boldsymbol{\omega})$ also obey the symmetry conditions. Therefore the image-domain representations of the odd parts are purely real.} that, if $F_{o1}(\boldsymbol{\omega})$ and $F_{o2}(\boldsymbol{\omega})$ are defined in this way, the corresponding image domain representations, $f_{o1}(\mathbf{x})$ and $f_{o2}(\mathbf{x})$, are purely real.

Comparing this definition to Equation (\ref{eqn:hilbertfrac}), it should be clear how this is a generalisation of the Hilbert transform (note that one can also include a minus sign in this definition, but it makes no difference as long as there is consistency). Each of the frequency components in each of the resulting odd parts is phase-shifted by $-\pi/2$, and also scaled such that the original amplitude is split between the two odd components.

It is also possible to express the Riesz transform as a convolution in the image domain, though we will not discuss it because the frequency domain representation is conceptually and computationally simpler (see \cite{Felsberg2000,Unser2009} for the relevant equations).

\subsection{Spherical Quadrature Filters}
\label{ssec:sqfs}

Just like in the 1D case, it is usually advantageous to filter the image with a 2D bandpass filter to introduce scale-selectivity before finding the monogenic signal. Just as in the 1D case, we can also think of this process as first finding the monogenic representation of the filters and then filtering the image with these filters.

In the 2D case, we start with a bandpass filter that is radially symmetric about the origin (`even') in both the frequency domain and image domain. We will again use the log-Gabor filter as an example, and note that a number of other filters are in common usage too (see \cite{Boukerroui2004}). The (even) 2D log-Gabor filter is given by:

\begin{equation}
G_e(\boldsymbol{\omega}) = \exp\left( -\frac{\left(\log\left(\frac{\lVert\boldsymbol{\omega}\rVert}{\omega_0}\right)\right)^2}{2\left(\log\left(\sigma_0\right)\right)^2}\right) .
\end{equation}

\begin{figure}
	\centering
	\includegraphics[width=\textwidth]{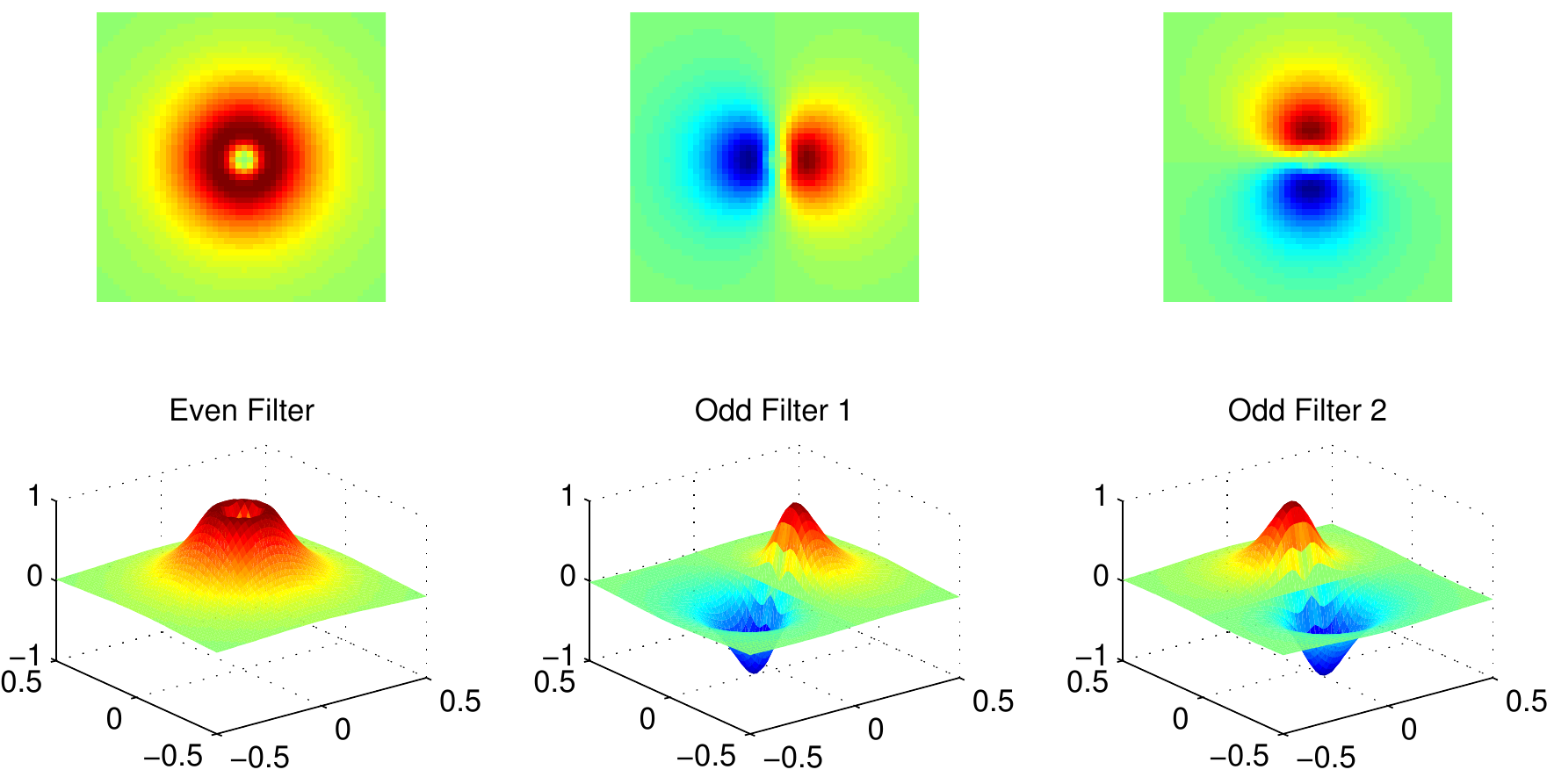}
	\caption{Frequency-domain representations of a set of log-Gabor spherical quadrature filters ($\sigma_0 = 0.5$).}
	\label{fig:filters2d_freq}
\end{figure}

\begin{figure}
	\centering
	\includegraphics[width=\textwidth]{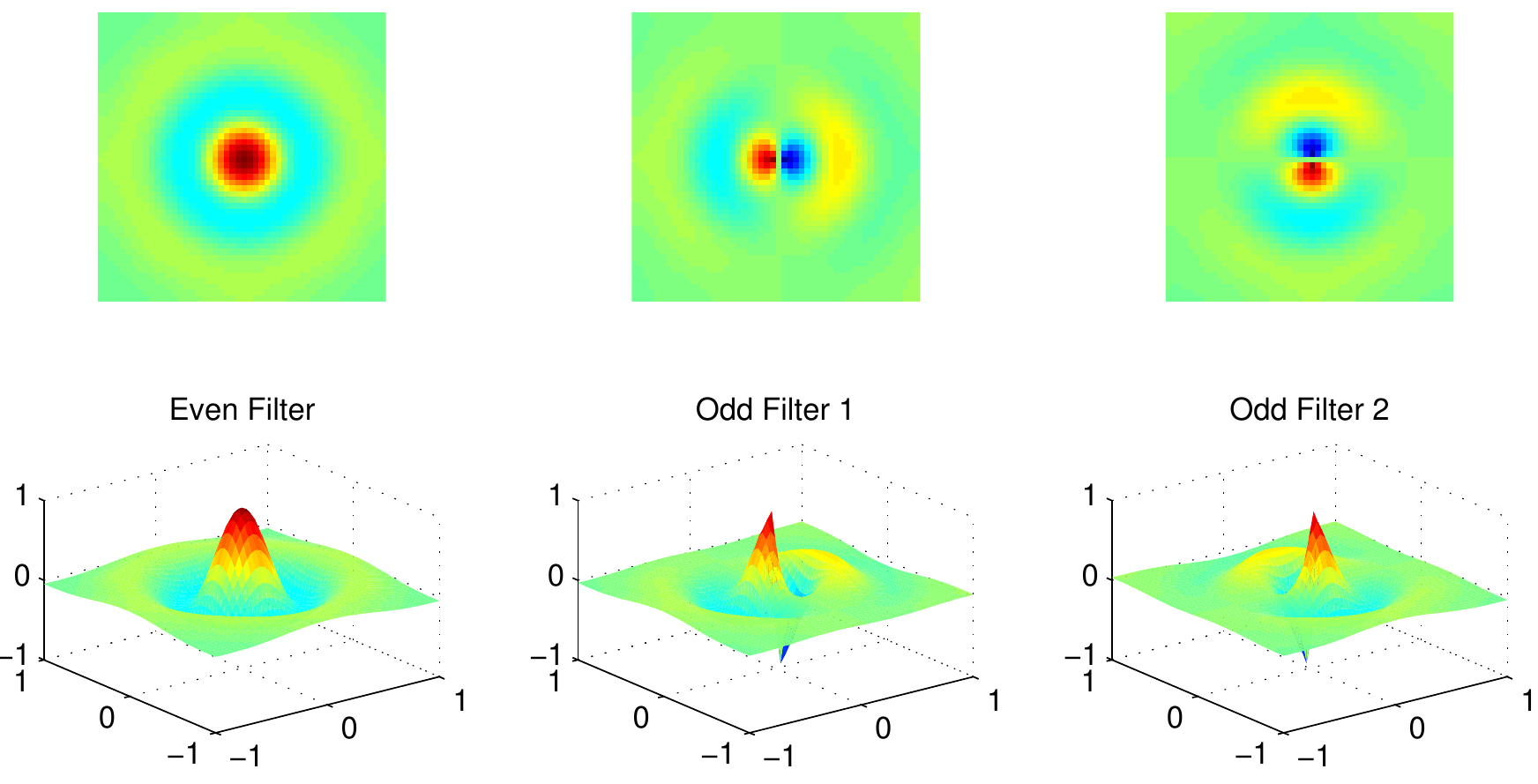}
	\caption{Image-domain representations (impulse responses) of a set of log-Gabor spherical quadrature filters ($\sigma_0 = 0.5$). Like the 1D case, there is no closed-form representation in the image domain, so these have been calculated numerically using the inverse FFT.}
	\label{fig:filters2d_im}
\end{figure}

We then form two odd parts of the filter, $G_{o1}(\boldsymbol{\omega})$ and $G_{o2}(\boldsymbol{\omega})$, using the Riesz transform, as in Equations (\ref{eqn:riesz1}) and (\ref{eqn:riesz2}). Each of the two resulting filters are odd-symmetric, with the axis of symmetry along the two image axes, see Figures \ref{fig:filters2d_freq} and \ref{fig:filters2d_im}. The resulting odd filter is isotropic because the combined amplitude, $\sqrt{G_{o1}(\boldsymbol{\omega})^2 + G_{o2}(\boldsymbol{\omega})^2} = G(\boldsymbol{\omega})$, is isotropic.

After filtering, we can write the monogenic signal as a combination of the three parts (one even, two odd) as a vector:

\begin{equation}
	\mathbf{f}_m(\mathbf{x}) = \begin{bmatrix} f_e(\mathbf{x}) \\ f_{o1}(\mathbf{x}) \\ f_{o2}(\mathbf{x}) \end{bmatrix}	 .
\end{equation}

\begin{figure}
	\centering
	\includegraphics[width=0.8\textwidth]{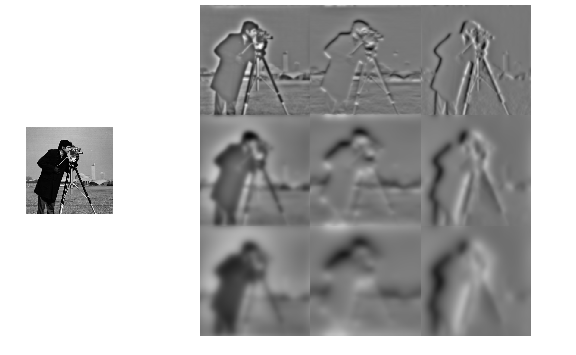}
	\caption{Cameraman test image (\textit{left}) and its monogenic signal (\textit{right}). In the right hand image: \textit{left column} $f_e(\mathbf{x})$, \textit{centre column} $f_{o1}(\mathbf{x})$, \textit{right column} $f_{o2}(\mathbf{x})$, \textit{top row} $\lambda_0 = 20$ pixels, \textit{centre row} $\lambda_0 = 60$ pixels,  \textit{bottom row} $\lambda_0 = 100$ pixels. Original image size $256 \times 256$ pixels.}
	\label{fig:monogeniccameraman}
\end{figure}

\begin{figure}
	\centering
	\includegraphics[width=0.8\textwidth]{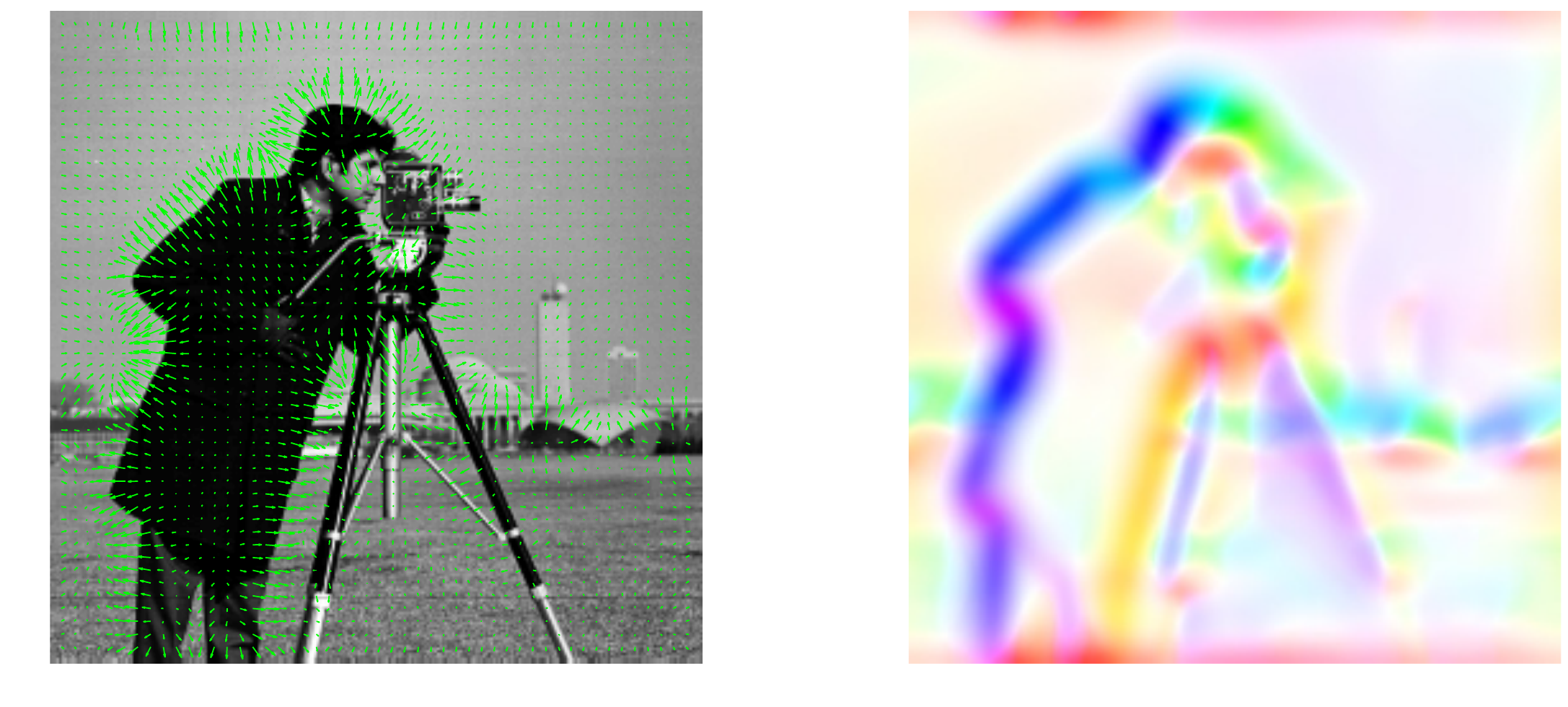}
	\caption{Alternative ways of visualising the odd filter response in the middle row of Figure \ref{fig:monogeniccameraman}: \textit{left} as a vector field superimposed over the original image, \textit{right} as a colour image with hue representing orientation and saturation representing magnitude.}
	\label{fig:oddmonogeniccameraman}
\end{figure}

See Figure \ref{fig:monogeniccameraman} for an example of the different components of the monogenic signal calculated at different wavelengths $\lambda_0 = 2\pi/\omega_0$. Notice how the two parts of the odd filter pick out horizontal and vertical changes in the image (recall the analogy with the smoothed gradient, which is very useful in 2D), and how structures of different sizes are picked out when filters of different wavelengths are used. Realising that the two parts of the odd filter are in fact part of one filter leads to more intuitive visualisations such as those in Figure \ref{fig:oddmonogeniccameraman}, which clearly demonstrate the odd filter picking out changes in the image and their directions.

It is also common to see the monogenic signal presented as the combination of one even part and \emph{one} odd part. Where this slightly misleading terminology occurs, the single odd part has been formed from the combined magnitude of the two odd parts. We will write the single odd part as $f_o(\mathbf{x})$ where:

\begin{equation}
	\label{eqn:combinedodd}
	f_o(\mathbf{x}) = \sqrt{f_{o1}(\mathbf{x})^2 + f_{o2}(\mathbf{x})^2} .
\end{equation}

Note that the two odd parts $f_{o1}(\mathbf{x})$ and $f_{o2}(\mathbf{x})$ must be calculated first and then combined to form the single part in the image domain (the magnitude cannot be found directly using a single filter). In some cases this notation is convenient, but beware that the orientation information is lost by performing this calculation.

\subsection{Local Phase, Amplitude, and Orientation in Images}
\label{ssec:imagephaseamplitude}

Equipped with the monogenic signal for 2D images, we are now able to extend the concepts of local phase and local amplitude to two dimensions. Recall how for 1D signals, we wrote the two parts of the analytic signal in a polar coordinate system and this gave us the amplitude and phase. In two dimensions, we have three components so we write them in a standard \emph{spherical polar} coordinate system, using the radius, elevation angle, and azimuthal angle.

\begin{figure}
	\centering
\tdplotsetmaincoords{60}{110}

\pgfmathsetmacro{\rvec}{.8}
\pgfmathsetmacro{\thetavec}{30}
\pgfmathsetmacro{\phivec}{60}

\begin{tikzpicture}[scale=5,tdplot_main_coords]

\coordinate (O) at (0,0,0);

\tdplotsetcoord{P}{\rvec}{\thetavec}{\phivec}


\draw[thick,->] (0,0,0) -- (1,0,0) node[anchor=north east]{$f_{o1}$};
\draw[thick,->] (0,0,0) -- (0,1,0) node[anchor=north west]{$f_{o2}$};
\draw[thick,->] (0,0,0) -- (0,0,1) node[anchor=south]{$f_e$};

\draw[-stealth,color=red] (O) -- node[below]{$A$} (P) node[anchor=south west]{$\mathbf{f}_m$};

\draw[dashed, color=red] (O) -- (Pxy) node[anchor=north west]{$\mathbf{f}_o$};
\draw[dashed, color=red] (P) -- (Pxy);

\tdplotdrawarc{(O)}{0.2}{0}{\phivec}{anchor=north}{$\theta$}

\tdplotsetthetaplanecoords{\phivec}

\tdplotdrawarc[tdplot_rotated_coords]{(0,0,0)}{0.5}{0}{\thetavec}{anchor=south west}{$\phi$}

\draw[dashed,tdplot_rotated_coords] (\rvec,0,0) arc (0:90:\rvec);
\draw[dashed] (\rvec,0,0) arc (0:90:\rvec);

\end{tikzpicture}
	\caption{Spherical polar representation representation of the components of the monogenic signal. The red arrow shows the monogenic signal at one point in an image. The angles $\phi$, representing the \emph{local phase}, and $\theta$, representing the \emph{local orientation}, are shown.}
	\label{fig:phasediag2d}
\end{figure}
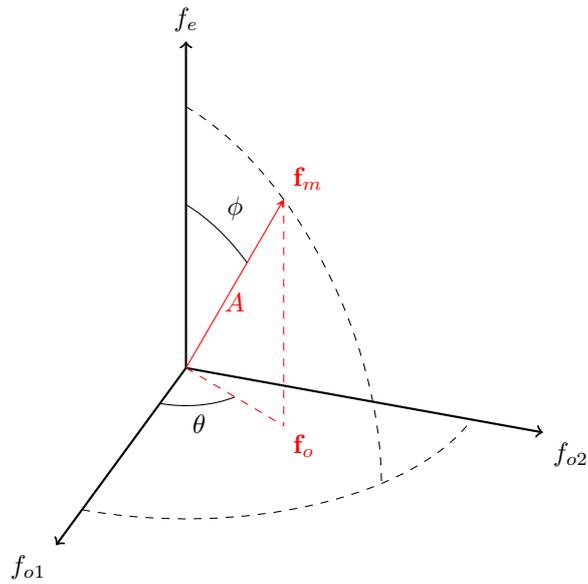

As before, the local amplitude is the radial part of the representation (see Figure \ref{fig:phasediag2d}):

\begin{align}
A(\mathbf{x}) &= \sqrt{f_e(\mathbf{x})^2 +  f_{o1}(\mathbf{x})^2 + f_{o2}(\mathbf{x})^2} \\
&= \sqrt{f_e(\mathbf{x})^2 +  f_o(\mathbf{x})^2} .
\end{align}

\noindent Additionally, some works refer to \emph{local energy}, which is simply the square of the local amplitude.

The local phase is found from the angle between the even part and the \emph{combined} odd part, and hence forms the elevation angle of the spherical polar representation:

\begin{equation}
	\phi(\mathbf{x}) = \arctan{\left(\frac{f_o(\mathbf{x})}{f_e(\mathbf{x})}\right)} .
\end{equation}

To complete the spherical polar representation we also need the \emph{local orientation}, $\theta$, which is simply the orientation of the odd filter response and represents the dominant direction in the image at that point (the orientation of the arrows in Figure \ref{fig:oddmonogeniccameraman}):

\begin{equation}
	\theta(\mathbf{x}) = \arctan{\left( \frac{f_{o2}(\mathbf{x})}{f_{o1}(\mathbf{x})}\right)} .
\end{equation}

The local orientation thus forms the azimuthal part of the spherical polar representation. The estimation of the local orientation can be unstable if the phase, $\phi$ is close to $0$, and so there some exist methods to estimate it more robustly (e.g.~\cite{Felsberg2000,Unser2009}). The isotropy of the monogenic filter means that, ignoring imperfection due to sampling, the local amplitude and local phase are unchanged by rotations of the input image while the local orientation rotates along with the input image.

We can interpret the image in the local area around a point as being approximated by a cosine wave with amplitude $A$ and phase $\phi$, and orientated on the image in direction $\theta$ \cite{Wietzke2009}. If we define a unit vector in the local orientation $\mathbf{n}(\mathbf{x}) = [\cos(\theta(\mathbf{x})),\sin(\theta(\mathbf{x}))]^T$ then around a point $\bar{\mathbf{x}}$ we have:

\begin{equation}
\label{eqn:2dphasemodel}
	f(\mathbf{x'}) \approx A(\bar{\mathbf{x}})\cos\left( \mathbf{x}'\cdot \mathbf{n}(\bar{\mathbf{x}}) + \phi(\bar{\mathbf{x}})\right) ,
\end{equation}

\noindent where $\mathbf{x}' = \mathbf{x} - \bar{\mathbf{x}}$. Notice that we are only modelling intensity variation in the direction of the local orientation. This means that the monogenic signal is useful for modelling image features such as edges and lines that have variation in one direction only (so called \emph{intrinsically 1D} or \emph{1iD} signals \cite{Wietzke2009}), but cannot model image features such as corners that have variation in two directions (\emph{intrinsically 2D} or \emph{2iD} signals). This is one important limitation of the monogenic signal.

Note also that there is a potential ambiguity here when defining the orientation and phase: for any given orientation and phase pair, we could always change the orientation to point in the opposite direction and change the sign of the phase to get another pair of values that are equally valid under the model in Equation (\ref{eqn:2dphasemodel}). To constrain this, in the 2D case the local phase can therefore only take values between $0$ and $\pi$ (this is a natural consequence of the fact that the combined odd filter $f_o(\mathbf{x})$ is always positive due to the definition in Equation (\ref{eqn:combinedodd})). Another way of looking at this is that the local orientation is defined to always point `uphill' towards areas of greater intensity, and the phase value is constrained to half the full range.

\begin{figure}
	\centering
	\includegraphics[width=\textwidth]{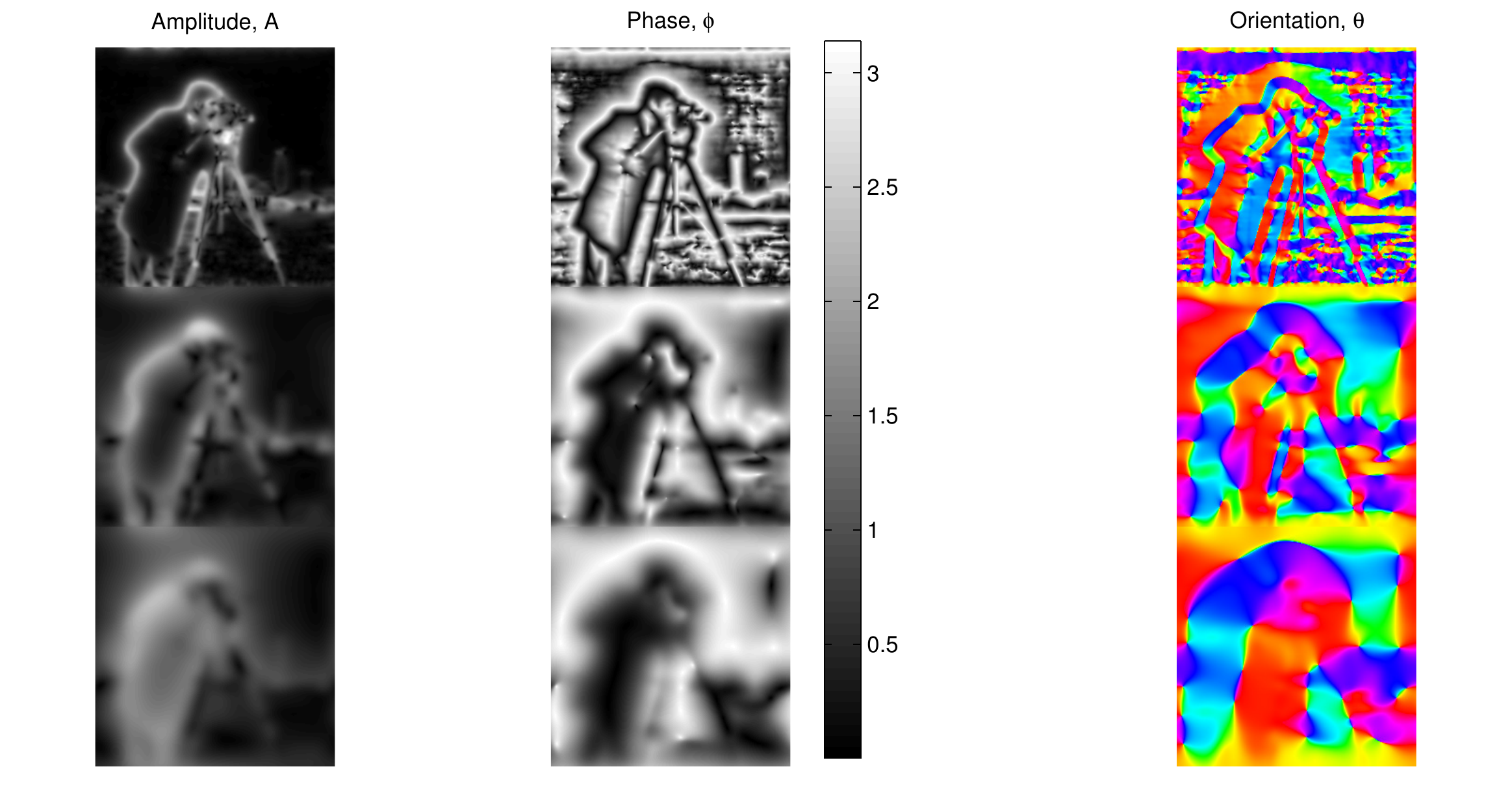}
	\caption{Local amplitude, phase and orientation images derived from the monogenic signals in Figure \ref{fig:monogeniccameraman}. The orientation image encodes angle as a hue on a colour wheel.}
	\label{fig:ampphaseori}
\end{figure}

The amplitude, phase and orientation images for the \textit{cameraman} image are shown in Figure \ref{fig:ampphaseori}.

\subsection{Complex Odd Filter Representation}
\label{ssec:complexoddfilter}

Whilst it is most accurate to represent the monogenic signal using quaternions (which is often undesirable in practical implementations), there are some advantages of representing just the odd part of the filter using a complex number. If we define a \textit{complex odd filter} in the frequency domain \cite{Unser2009} as:

\begin{align}
	G_{oc}(\boldsymbol{\omega}) &= G_{o1}(\boldsymbol{\omega}) + \imagi G_{o2}(\boldsymbol{\omega}) \\
	 &= \imagi \frac{\omega_x}{\lVert\boldsymbol{\omega}\rVert} G_e(\boldsymbol{\omega}) + \imagi \left( \imagi \frac{\omega_y}{\lVert\boldsymbol{\omega}\rVert} G_e(\boldsymbol{\omega})\right) \\
	 &= \frac{\imagi\omega_x - \omega_y}{\lVert\boldsymbol{\omega}\rVert}G_e(\boldsymbol{\omega}) .
\end{align}

\noindent By linearity of the Fourier transform, in the image domain this gives us an impulse response of

\begin{equation}
	g_{oc}(\mathbf{x}) = g_{o1}(\mathbf{x}) + \imagi g_{o2}(\mathbf{x}) ,
\end{equation}

\noindent i.e.\ a complex-valued filter whose real and imaginary parts are the two parts of the odd filter. Convolving this complex filter with an image then gives us an image $f_{oc}(\mathbf{x})$ whose real and imaginary parts are the two odd parts of the image's monogenic signal. This is convenient because the local orientation is intuitively recovered as the argument of the complex number\footnote{This puts the complex odd filter neatly within the framework of rotationally \emph{equivariant} filters that are used for rotation-invariant description in several works, e.g.~\cite{Liu2014}}, and the amplitude of the complex number is the magnitude of the combined odd filter:

\begin{align}
\theta(\mathbf{x}) &= \arg\left(f_{oc}(\mathbf{x})\right)\\
f_o(\mathbf{x}) &= \lvert f_{oc}(\mathbf{x}) \rvert \\
A(\mathbf{x}) &= \sqrt{f_e(\mathbf{x})^2 + \lvert f_{oc}(\mathbf{x}) \rvert ^2}\\
\phi(\mathbf{x}) &= \arctan\left(\frac{\lvert f_{oc}(\mathbf{x}) \rvert}{f_e(\mathbf{x})}\right) .
\end{align}

Using a complex odd filter also has computational advantages. Since performing an inverse fast Fourier transform is inherently an operation involving complex numbers, it is generally no more computationally expensive to perform the inverse transform $f_{oc}(\mathbf{x}) = \mathcal{F}^{-1}\lbrace F_{oc}(\boldsymbol{\omega}) \rbrace$ than it is to perform either of the filters individually $f_{o1}(\mathbf{x}) = \mathcal{F}^{-1}\lbrace F_{o1}(\boldsymbol{\omega}) \rbrace$ or $f_{o2}(\mathbf{x}) = \mathcal{F}^{-1}\lbrace F_{o2}(\boldsymbol{\omega}) \rbrace$. Therefore, by implementing a single complex filter rather than two separate filters, we save ourselves one FFT operation. This is particularly important where processing time is important, for example in video applications.

\begin{figure}
	\centering
	\includegraphics[width=0.7\textwidth]{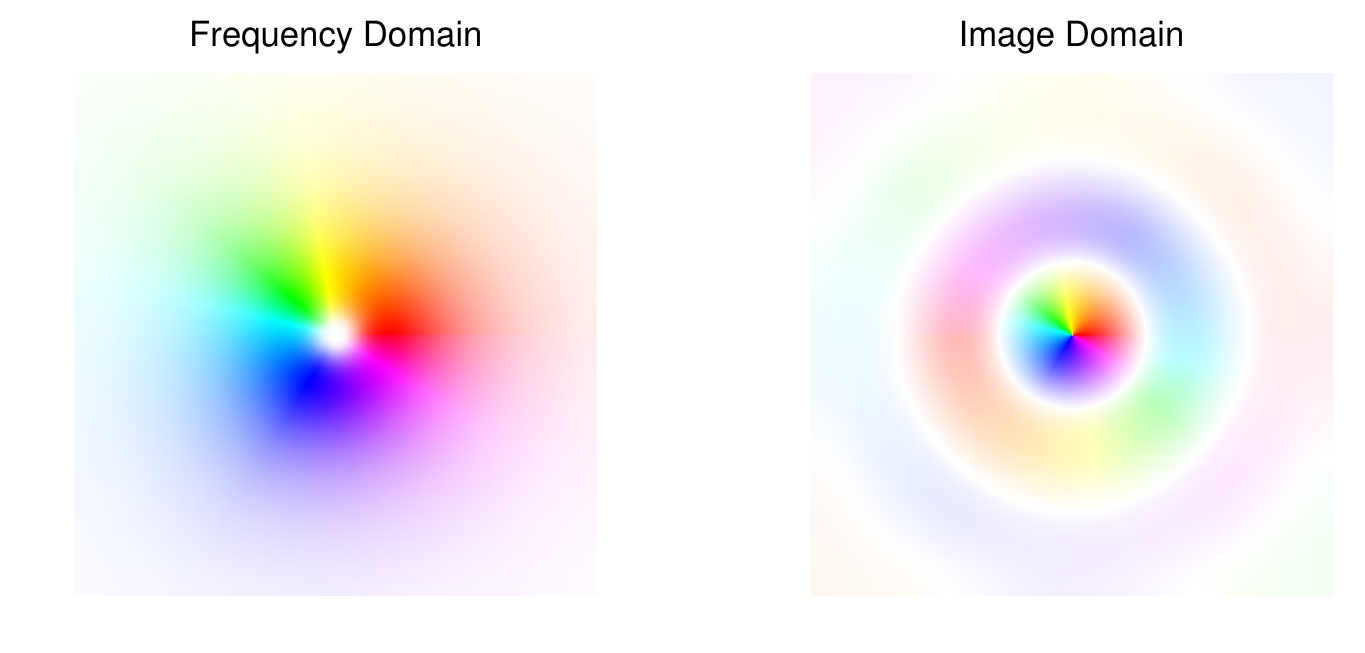}
	\caption{Representations of complex valued filters: (\textit{left}) the odd filters from Figure \ref{fig:filters2d_freq} shown in the frequency domain, (\textit{right}) the odd filters from Figure \ref{fig:filters2d_im} shown in the image domain. In each case, the hue represents the argument of the complex value, and the saturation represents the magnitude.}
	\label{fig:filters2d_comp}
\end{figure}

Figure \ref{fig:filters2d_comp} shows the complex-valued alternatives to the odd filter pairs shown in previous examples. From this representation it is easier to see that the combined effect of the odd filters is isotropic. Because the argument of the complex number is the local orientation, if we rotate the input image by an angle, $\psi$, the resulting complex odd filter response, $f_{oc}(\mathbf{x})$, is simply rotated by multiplication by the unit complex number $e^{\imagi\psi}$.

\section{Quantities Derived from the Monogenic Signal}
\label{sec:derived}

We have now introduced the monogenic signal for image analysis. This section will introduce some important quantities that may be derived from the monogenic signal. Beware that the quantities discussed here pre-date the development of the monogenic signal, so many earlier texts used sets of quadrature filters at a range of orientations to calculate the local amplitude, phase, and orientation measures.

\subsection{Phase Symmetry and Asymmetry}
\label{ssec:sym}

It is often useful to measure the extent to which the structure around a point in an image is locally \emph{symmetric} or \emph{asymmetric}. For a signal to be symmetric, all the constituent frequency components must be symmetric, i.e.\ all frequencies must have a phase of $0$ or $\pi$ or equivalently be pure cosine waves. Conversely all components of an asymmetric signal must have phases of $\pi/2$ or $3\pi/2$ or equivalently be pure sine waves.

One way to quantify symmetry therefore is to find the difference between the sine and cosine of the local phase. Kovesi's feature symmetry $S$ and asymmetry $R$ measures \cite{Kovesi1997} are based on this principle:

\begin{align}
\label{eqn:symsincos}
S(\mathbf{x}) = \frac{\lfloor\,\lvert \cos(\phi(\mathbf{x})) \rvert - \lvert \sin(\phi(\mathbf{x})) \rvert - T \,\rfloor}{A(\mathbf{x}) + \epsilon} \\
\label{eqn:asymsincos}
R(\mathbf{x}) =  \frac{ \lfloor \,\lvert  \sin(\phi(\mathbf{x})) \rvert - \lvert \cos(\phi(\mathbf{x}))  \rvert - T \,\rfloor}{A(\mathbf{x}) + \epsilon} ,
\end{align}

\noindent where $T$ is a threshold that sets the sensitivity of the measure, $\lfloor \cdot \rfloor$ is an unconventional shorthand meaning that values less than zero are replaced with zero, and $\epsilon$ is a small number that avoids division by zero. Normalising by the local amplitude gives a measure of symmetry that is independent of amplitude (i.e.\ a purely structural measure) and lies in the range $0 \leq S(\mathbf{x}) < 1$ and $0 \leq R(\mathbf{x}) < 1$.  A large number of pixels in the image have symmetry/asymmetry values of $0$ due to the thresholding.

The definitions of symmetry and asymmetry in equations (\ref{eqn:symsincos}) and (\ref{eqn:asymsincos}) above are equivalent to the following, written in terms of parts of the monogenic signal, using the definition of the combined (directionless) odd filter in Equation (\ref{eqn:combinedodd}):

\begin{align}
\label{eqn:sym}
S(\mathbf{x}) = \frac{\lfloor\,\lvert f_e(\mathbf{x}) \rvert - \lvert f_o(\mathbf{x}) \rvert - T \,\rfloor}{A(\mathbf{x}) + \epsilon} \\
\label{eqn:asym}
R(\mathbf{x}) =  \frac{ \lfloor \,\lvert f_o(\mathbf{x}) \rvert - \lvert f_e(\mathbf{x}) \rvert - T \,\rfloor}{A(\mathbf{x}) + \epsilon} .
\end{align}

This is intuitive because the magnitude of the response of a near-symmetric function to the even-symmetric filter $g_e(\mathbf{x})$ will be large and the magnitude of the response to the odd-symmetric filter $g_{o}(\mathbf{x})$ will be small. The opposite holds for functions that are nearly asymmetric.

Equations (\ref{eqn:sym}) and (\ref{eqn:asym}) measure scale at just one scale in the image (the scale of the bandpass filter used to calculate the monogenic signal). Depending on the application, it is often more useful to accumulate results at a number of scales. If  we denote properties calculated at a wavelength $\lambda$ using a subscript $i$ then a multiscale measure of symmetry is given by

\begin{align}
\label{eqn:symmultiscale}
S(\mathbf{x}) = \sum_{i}\frac{\lfloor\,\lvert f_{e,\lambda_i}(\mathbf{x}) \rvert - \lvert f_{o,\lambda_i}(\mathbf{x}) \rvert - T \,\rfloor}{A_{\lambda_i}(\mathbf{x}) + \epsilon} \\
\label{eqn:asymmultiscale}
R(\mathbf{x}) =  \sum_i \frac{ \lfloor \,\lvert  f_{o,\lambda_i}(\mathbf{x}) \rvert - \lvert f_{e,\lambda_i}(\mathbf{x}) \rvert - T \, \rfloor}{A_{\lambda_i}(\mathbf{x}) + \epsilon} ,
\end{align}

\noindent where $\lbrace\lambda_i\rbrace$ are a set of filter centre-wavelengths chosen to suit the application.

Note that there are other ways of obtaining a phase-based description of images, such as using a set of filters with different orientations or a steerable filters, and these have frequently been used in the past instead of the monogenic signal \cite{Kovesi1997}.

\begin{figure}
	\centering
		\includegraphics[width=\textwidth]{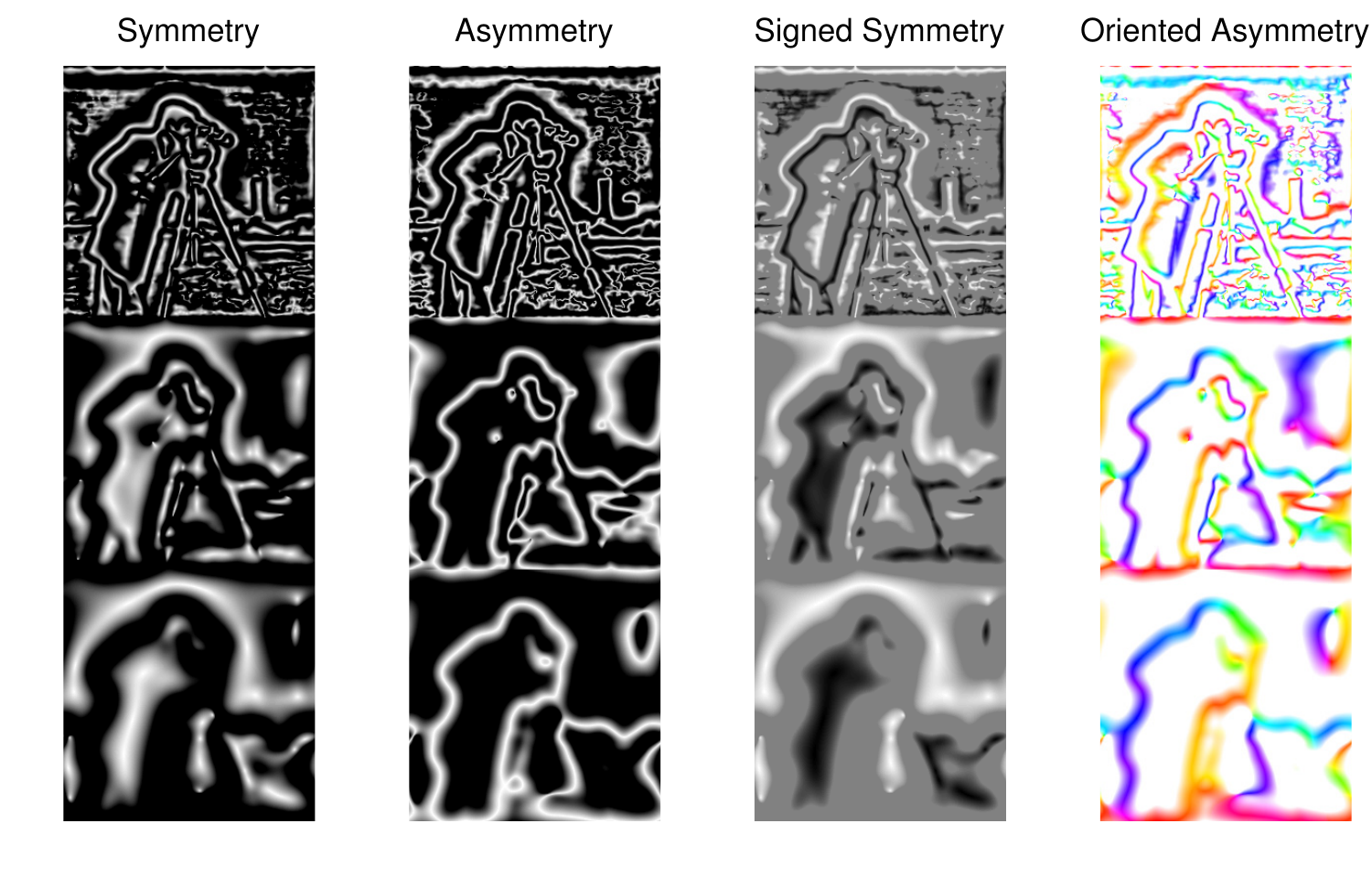}
		\caption{\textit{First column} symmetry and \textit{second column} asymmetry calculated for the cameraman test image using the same filter parameters as the corresponding rows in Figure \ref{fig:monogeniccameraman} and a threshold $T = 0.18$. Values are displayed from 0 (black) to 1 (white). \textit{Third column} signed symmetry and \textit{fourth column} oriented asymmetry calculated for the same images and parameters. Signed symmetry is shown on a scale of -1 (black) to 1 (white). Oriented asymmetry is shown as a colour image with hue representing orientation and saturation representing magnitude.}
		\label{fig:symcameraman}
\end{figure}

\begin{figure}
	\centering
	\includegraphics[width=\textwidth]{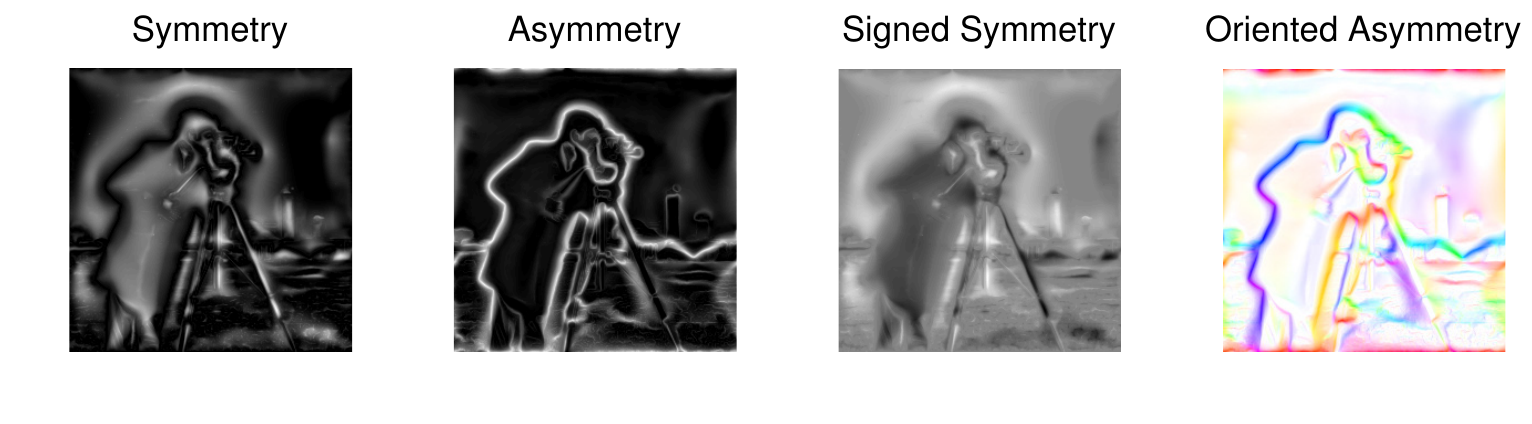}
	\caption{Multiscale symmetry and asymmetry for the cameraman test image using a set of log-Gabor filters with centre-wavelengths in the range 5 to 100 pixels and with a spacing of 5 pixels  (i.e.\ $\lbrace5, 10, 15,\ldots,100\rbrace$) and a shape parameter of $\sigma_0 = 0.3$. The threshold used was and a threshold $T = 0.18$. Image interpretation is as in Figure \ref{fig:symcameraman}.}
	\label{fig:multisymcameraman}
\end{figure}

We return to the cameraman test image for an illustration of feature symmetry and asymmetry (see Figure \ref{fig:symcameraman}). We can see that feature symmetry picks out `blobs' (regions of similar appearance) at the scale of interest whereas feature asymmetry picks edge-like features that form the boundaries between them. Figure \ref{fig:multisymcameraman} shows an example of the multi-scale measure on the same image. Notice how multi-scale feature asymmetry is particularly helpful for detecting useful boundaries.

\subsection{Retaining Orientation in Symmetry and Asymmetry}
\label{ssec:orsym}

The potential usefulness of feature symmetry and asymmetry for detecting edges and structures of interest should be apparent from Figures \ref{fig:symcameraman} and \ref{fig:multisymcameraman}. Notice, however, that when we found a value for symmetry and asymmetry we always took the magnitude of the monogenic signal parts (Equations (\ref{eqn:sym}) and (\ref{eqn:asym})). The result is that we throw away some information that may be useful to us, depending on the application.

In the case of feature symmetry, using the magnitude of the monogenic even part means that we are discarding information about whether the pixel sits in a \emph{peak} or a \emph{trough}. This can be rectified by retaining the sign of the even part in Equation (\ref{eqn:sym}) by simply multiplying by the signum function:

\begin{align}
\label{eqn:signedsym}
\hat{S}(\mathbf{x}) &= \frac{\lfloor\,\lvert f_e(\mathbf{x}) \rvert - \lvert f_o(\mathbf{x})\, \rvert - T \rfloor}{A(\mathbf{x}) + \epsilon} \cdot \sgn(f_e(\mathbf{x}))\\
&= S(\mathbf{x}) \sgn(f_e(\mathbf{x})) \\
&= S(\mathbf{x})\frac{f_e(\mathbf{x})}{\lvert f_e(\mathbf{x}) \rvert}, \quad f_e(\mathbf{x})\neq 0 .
\end{align}

This gives a \emph{signed feature symmetry} value in the range $-1 \leq \hat{S}(\mathbf{x}) < 1$.
To my knowledge, this has not been explicitly described before except by my recent work \cite{Bridge2015}. However, some authors may have used a similar technique in earlier work, for example inspection of figures in \cite{Rahmatullah2012} suggests that this technique may have been used.

In the case of feature asymmetry, by taking the magnitude of the monogenic odd part in Equation (\ref{eqn:asym}), we are throwing away the orientation of the edge. Edge orientation is particularly useful in many contexts, so this seems like something of an oversight. We can rectify this in a similar way, by multiplying by a unit complex number in the direction of the odd part found from the complex formulation of the odd filter. This gives a complex number $\hat{R}(\mathbf{x})$ that lies within the unit disc (i.e.\ $0 \leq \lvert \hat{R}(\mathbf{x})\rvert < 1$) quantifying both the magnitude and direction of the asymmetry (one could equally use a vector formulation, but a complex formulation is convenient when working with a complex odd filter definition):

\begin{align}
\label{eqn:orasym}
\hat{R}(\mathbf{x}) &=  \frac{ \lfloor \, \lvert f_o(\mathbf{x}) \rvert - \lvert f_e(\mathbf{x}) \rvert - T \,\rfloor}{A(\mathbf{x}) + \epsilon} \cdot \frac{f_{oc}(\mathbf{x})}{\lvert f_{oc}(\mathbf{x})\rvert} \\
&= R(\mathbf{x})\frac{f_{oc}(\mathbf{x})}{\lvert f_{oc}(\mathbf{x})\rvert} .
\end{align}

Figures \ref{fig:symcameraman} and \ref{fig:multisymcameraman} compare signed feature symmetry with feature, and oriented feature asymmetry with feature asymmetry for the cameraman test image.

\subsection{Phase Congruency}
\label{ssec:congruency}

\emph{Phase congruency} is a further, related measure for feature detection in images. Where feature symmetry and asymmetry measure how close the local phase value is to 0 or $\pi$ (symmetry) and $\pi/2$ (asymmetry), phase congruency measures how similar phase values at different scales are to \emph{each other}. The intuition here is that interesting features appear where the phase values at different scales are the same. We will not discuss phase congruency in much detail as it has not been used much in ultrasound analysis. However it is widely used to detect edges and features in natural images. We refer the interested reader to Koversi's work \cite{Kovesi2003} for more details.

\section{Application to Ultrasound Image Analysis}
\label{sec:us}

In this section we shall briefly summarise the utility of the monogenic signal and related measures in analysis of ultrasound images and refer the reader to the literature for more details on the specific techniques.

The monogenic signal, feature symmetry/asymmetry, and phase congruency have proven useful in ultrasound analysis tasks where more traditional techniques have failed. There are two major important reasons for this:

\begin{itemize}
\item By tuning the frequency of the bandpass filters, it is possible to pick out features at the desired scale and ignore small scale speckle and large scale variations such as shadowing artefacts if desired.
\item Phase-based measures are robust to the contrast variation that can occur due to different imaging parameters and other effects such as shadowing and enhancement.
\end{itemize}

\noindent It has found use in a number of applications including:

\begin{itemize}
\item Boundary detection: Traditional edge detection methods tend to perform poorly on the indistinct edges in ultrasound images, and are confounded by speckle. Feature asymmetry has proven effective in detecting boundaries in both 2D. For example, Mulet-Parada and Noble~\cite{MuletParada2000} used feature asymmetry to detect boundaries in echocardiograms, though they used a set of oriented real-valued filters rather than the monogenic signal. Rajpoot et al.~\cite{Rajpoot2008} used the monogenic signal for the same purpose, and this was also extended to 3D ultrasound images \cite{Rajpoot2009,Stebbing2013}.
\item Motion estimation: Traditional `optical flow' techniques tend to be based on the assumption that the brightness of corresponding image points does not change as the image moves. Felsberg \cite{Felsberg2006} instead assumed the constancy of local phase between frames to derive an alternative method. Alessandrini et al.~\cite{Alessandrini2013} built on this method to create a motion estimation algorithm for cardiac ultrasound and cardiac MR imaging.
\item Registration: Similar to motion estimation, Grau et al.~\cite{Grau2006} used a similarity measure based on local phase rather than intensity for the alignment of 3D echocardiography imagery and find it to be more robust than intensity-based methods.
\item Compounding: Compounding is the process of `fusing' multiple aligned images to create a combined image that has the `best' parts of all the original images. This usually involves using some sort of `feature importance' measure to determine which parts of the final image should be taken from which input image. Grau and Noble \cite{Grau2005} used a measure based on phase congruency for this purpose.
\item Recognition and detection: The monogenic signal has been employed in diverse ways towards recognition in ultrasound images. Rahmatullah et al.~\cite{Rahmatullah2012} and Patwardhan \cite{Patwardhan2012} used feature symmetry to find candidate `blobs' for detection of the stomach and umbilical vein in ultrasound. Maraci et al.~\cite{Maraci2014} used histograms of keypoints based on dense SIFT descriptors calculated on a local phase image in order to recognise different views in an ultrasound video of the fetus. Bridge and Noble \cite{Bridge2015} incorporated various monogenic signal-based image representations into a rotation invariant sliding window object detection framework in order to localise the fetal heart in ultrasound videos.
\end{itemize}

However, beware the following limitations, as the monogenic signal is not appropriate for all applications in ultrasound:

\begin{itemize}
\item Information is lost by using the monogenic signal due to the bandpass filtering process. This may mean, for example, that speckle information and absolute intensity information is lost. Though this is often useful to provide invariance to contrast and robustness to speckle, it is important to consider whether this is appropriate for a particular application. Particularly in the case of applications that make use of learning techniques, it is important to realise that using monogenic features is forcing the algorithm to use a certain low-level representation that may be suboptimal.
\item As described in \S\ref{ssec:imagephaseamplitude}, the monogenic signal assumes that the image is locally intrinsically one dimensional. This means that it is not capable of analysing 2D features such as corners or curvature.
\item In the language of computer vision, the monogenic signal and derived features could be considered `hand-crafted' features, based on mathematical principles rather than being optimised for the task at hand. Over the last few years, the general computer vision community has been moving away from such hand-crafted features in favour of learning task-specific image representations with techniques such as convolutional neural networks, which have achieved the recent state-of-the-art results on detection and recognition tasks. Such learned features have yet to be thoroughly investigated in the domain of ultrasound imaging, perhaps in part because of the large amount of training examples typically needed to train them.
\end{itemize}

\section{Extensions and Generalisations}
\label{sec:extensions}

Here we briefly summarise some extensions to the theory presented so far that may be of interest to those working with the monogenic signal.

\begin{itemize}
\item Scale Space: Felsberg and Sommer \cite{Felsberg2004} developed a mathematical framework for conducting scale-space analysis using phase-based processing, based on the link between the monogenic signal and the Possion bandpass filter.
\item Wavelets: Unser et al.~\cite{Unser2009} have developed a multi-scale wavelet analysis based on the Riesz transform that is essentially a monogenic wavelet decomposition acting in the frequency domain. A fast image domain approximation has been developed by Wadhwa et al.~\cite{Wadhwa2014} and has been used to provide amplification of imperceptible motion in video.
\item 2D Analytic Signal: It is possible to extend the monogenic signal to enable it to characterise image regions that are intrinsically 2D (i.e.\ have curvature in the second dimension such as a corner). Wietzke and colleagues \cite{Wietzke2009,Wietzke2010,Wietzke2008} have achieved this using additional filters found from the even filter using a \emph{second-order} Riesz transform and call the result (perhaps confusingly) the \emph{2D analytic signal}. This has been used on radio frequency ultrasound data in order to improve the envelope detection method \cite{Wachinger2012a}.
\end{itemize}

\section{Acknowledgements}

I would like to thank Ana Namburete and Thomas Nketia for the initial idea to write this document, and Davis Vigneault and Ruobing Huang for their comments on how to improve early drafts.

\bibliographystyle{unsrt}
\bibliography{monogenic.bib}

\end{document}